# Recovering Dense Tissue Multispectral Signal from *in vivo* RGB Images

Jianyu Lin[1,2], Neil T. Clancy[1,3], Daniel S. Elson[1,3]

[1]The Hamlyn Centre for Robotic Surgery, Imperial College London, London, SW7 2AZ, UK
[2]Department of Computing, Imperial College London, London, SW7 2AZ, UK
[3]Department of Surgery and Cancer, Imperial College London, London, SW7 2AZ, UK
j.lin12 @imperial.ac.uk

## INTRODUCTION

Hyperspectral/multispectral imaging (HSI/MSI) contains rich information clinical applications, such as 1) narrow band imaging for vascular visualisation; 2) oxygen saturation for intraoperative perfusion monitoring and clinical decision making [1]; 3) tissue classification and identification of pathology [2]. The current systems which provide pixel-level HSI/MSI signal can be generally divided into two types: spatial scanning and spectral scanning. However, the trade-off between spatial/spectral resolution, the acquisition time, and the hardware complexity hampers implementation in real-world applications, especially intra-operatively. Acquiring high resolution images in real-time is important for HSI/MSI in intra-operative imaging, to alleviate the side effect caused by breathing, heartbeat, and other sources of motion. Therefore, we developed an algorithm to recover a pixel-level MSI stack using only the captured snapshot RGB images from a normal camera. We refer to this technique as "super-spectral-resolution". The proposed method enables recovery of pixel-level-dense MSI signals with 24 spectral bands at ~11 frames per second (FPS) on a GPU. Multispectral data captured from porcine bowel and sheep/rabbit uteri *in vivo* has been used for training, and the algorithm has been validated using unseen *in vivo* animal experiments.

## MATERIALS AND METHODS

Super-resolution, which estimates high resolution (HR) images from low resolution (LR) counterparts, is a highly ill-posed problem, since one LR image could be matched to many possible HR images. This problem is solvable only under certain constraints. In this work, we made the following assumptions: 1) optimised matching can be found by learning the training set which contains similar information to the unseen test set; 2) the HR information from the HR images could be recovered from its LR correspondences; 3) there is no correlation between adjacent pixels.

Recently, deep learning techniques including convolutional neural networks (CNNs) has been applied on image super-resolution, and proved to provide promising results [3, 4]. In this work, similar strategies were adopted to upscale an RGB image which contains 3 channels ($M \times N \times 3$ matrix), to an MSI stack with 24 channels ($M \times N \times 24$ matrix).

The proposed model consists of two phases (Fig. 1):

- **The upscaling phase:** Three 3D deconvolutional layers followed by one convolutional layer were piled together to transform the input from $M \times N \times 3$ to $M \times N \times 24$
- **The high resolution extraction (HRE) phase:** A residual network block was used to extract and combine the high frequency (HF) with low frequency information from the upscaling phase's output.

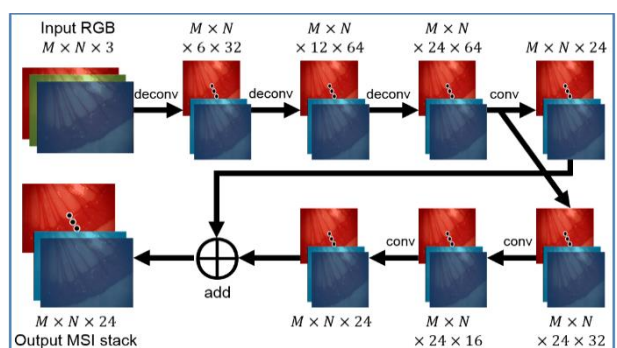

**Fig. 1** The prediction model. The input is a RGB image with a user-defined dimension ($M \times N$); the output is an MSI stack with the same spatial dimension and 24 channels.

MSI stacks collected from porcine bowel and sheep/rabbit uteri *in vivo* with a liquid crystal tunable filter (LCTF) endoscopic imager were used for training and validation [2]. The MSI stacks were collected at 10 nm spectral interval, in the range of 460-720 nm. A non-rigid registration method was used to eliminate the small displacement between MSI slices at different spectral bands [5], generating the MSI stacks as the ground truth for training. Synthetic RGB images were simulated from these MSI stacks, with the transmission spectrum measured from a normal RGB camera (Thorlabs DCU223C). The training set contains the 243 MSI stacks (augmented from 50 porcine bowel, 21 rabbit uterus, and 10 sheep uterus). During training, based on assumption no. 3, every pixel was trained independently, with convolutional kernels expanding along the spectral dimension only. Then the learnt weights were used to predict a whole MSI stack from input RGB images, by only changing the input data dimension from ($1 \times 1 \times 3$) in the training network to ($256 \times 192 \times 3$) in the prediction counterpart. Both the training and prediction were applied with Tensorflow [6]. Predicting the MSI stack from an RGB image (256 × 192) cost ~90 ms on a PC (CPU: i7-3770; GPU: NVIDA GTX TITAN X).

**RESULTS**

The proposed algorithm was evaluated for two outputs: the overall and inter-class prediction. The peak signal-to-noise ratio (PSNR) was used to quantify the MSI stack prediction accuracy.

Firstly, all the *in vivo* data from different sources were mixed and underwent a 5-fold cross validation. The PSNR value for different spectral bands were evaluated by comparing the calculation results and the ground truth. The average and standard deviation of PSNR were plotted alongside the RGB camera transmission spectra in Fig. 2 (upper). It can be found that PSNR is relatively high (>28) for most spectral bands, but drops in the blue/green/red spectra overlapping areas. We also demonstrates the accuracy of MSI estimation intuitively, by comparing the estimated MSI signal with the ground truth, at different locations on a rabbit uterus (Fig. 2 (lower)). As is known, HSI/MSI signal can be used to calculate the oxygen saturation, by fitting it using a linear combination of absorption spectra of oxy/deoxy-haemoglobian, as is shown in Fig. 2 (upper).

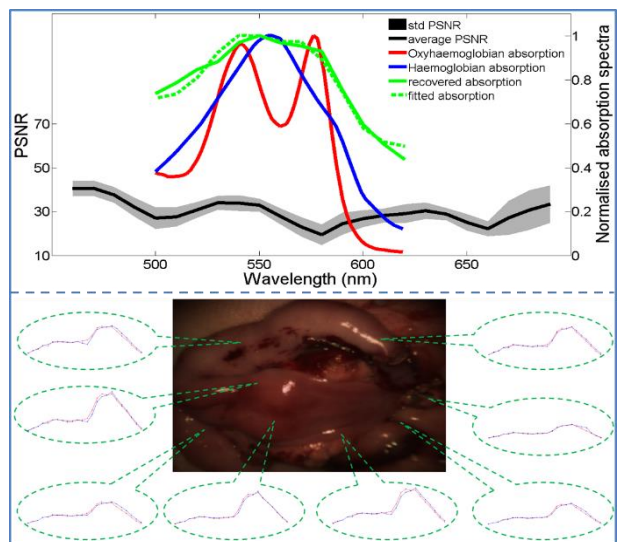

**Fig. 2** Upper: average (black line) and standard deviation (yellow shadow) of the PSNR values for all wavelengths; the normalised absorption spectra for oxy/deoxy-haemoglobian (red and blue line), together with the estimated and fitted MSI signal (green and green dash line); Lower: comparison between the estimated (blue line) and ground truth (red line) MSI signal at different locations on a rabbit uterus.

The inter-class prediction was also validated. The models trained on individual classes were used to predict the MSI stacks on other classes, and peak signal-to-noise ratio (PSNR) was adopted to evaluate the results (Table 1). Promising results (PSNR > 30) were achieved on sheep and rabbit uteri, while the training using pig bowel led to the worst outcome (PSNR ≈ 26). This is mainly due to the more frequent presence of specular reflections, and larger error in generating the MSI ground truth due to imperfect registration. This result shows the potential of transfer learning to use this technique to estimate MSI of tissues with limited available samples by training on other similar datasets, e.g., prediction of human tissue MSI using the training result from *in vivo* animal data.

**Table. 1** PSNR for inter-class MSI prediction. Data from three sources were used for the validation: 1) pig bowel (PB); 2) sheep uterus (SU); 3) rabbit uterus (RU).

| Test / Train | PB | SU | RU |
|---|---|---|---|
| PB | **25.75** | 27.47 | 28.80 |
| SU | 24.80 | **32.49** | 32.34 |
| RU | 25.08 | 31.83 | **32.96** |

**DISCUSSION**

In this work, we developed a CNN based model to recover dense pixel level MSI signals from RGB images, in real-time. As the HSI/MSI signal is important for many applications such as narrow-band imaging, tissue classification, and tumor detection, we believe that this technique can benefit these applications. It overcomes the problem of long acquisition time for normal HSI scanning systems and the need for separate specialized imaging equipment, providing promising MSI estimation with only slight accuracy loss in most spectral bands. In future work the research will focus on: 1) the validation of the current algorithm on more *in vivo* data, especially human tissue; 2) the extension of both the algorithm and the hardware to achieve higher accuracy.